\documentclass[letterpaper]{article}
\usepackage{iccc}

\usepackage{graphicx}
\usepackage{times}
\usepackage{helvet}
\usepackage{courier}
\usepackage[hyphens]{url}
\title{Predicting User Perception of Move Brilliance in Chess }
\author{Kamron Zaidi and Michael Guerzhoy\\
University of Toronto\\
kamron.zaidi@mail.utoronto.ca, guerzhoy@cs.toronto.edu}

\begin{document} 
\maketitle

\begin{abstract}
\begin{quote}
{\em AI research in chess has been primarily focused on
producing stronger agents that can maximize the probability
of winning. However, there is another aspect to chess that
has largely gone unexamined: its aesthetic appeal.
Specifically, there exists a category of chess moves called
``brilliant" moves. These moves are appreciated and admired
by players for their high intellectual aesthetics. We demonstrate the first system for classifying chess moves as brilliant. The system uses a neural network, using the output of a chess engine as well as features that describe the shape of the game tree. The
system achieves an accuracy of 79\% (with 50\% base-rate), a PPV of 83\%, and an NPV of 75\%. We demonstrate that what humans perceive as ``brilliant" moves is not merely the best possible move. We show that a move is more likely to be predicted as brilliant, all things being equal, if a weaker engine considers it lower-quality (for the same rating by a stronger engine). Our system opens the avenues for computer chess engines to (appear to) display human-like brilliance, and, hence, creativity.}
\end{quote}
\end{abstract}

\section{Introduction}
Superhuman chess AI has made significant advances in recent decades. In this paper, we focus on human-like play, and specifically play that is aesthetically appealing to humans, in the sense that it's perceived as brilliant and/or creative\footnote{``Brilliant," ``aesthetically appealing," and ``creative" can philosophically be treated as distinct. However, in the context of chess, those are closely related.}. Chess can be
thought of as an intellectual art, as it allows for the perception
of beauty~\cite{Osborne1964ChessAesthetics}. Moves termed as ``brilliant" are highly
regarded by players for a variety of factors relating to the
aesthetic appreciation of the move. The practice of awarding
brilliancy prizes at chess tournaments for the most beautiful
moves/games gives credence to the aesthetic nature of chess and the
value placed on it by chess players~\cite{Humble1993ChessArt}. Aesthetics also appears
outside the context of full games, such as in chess puzzles, where
aesthetics is even more central to the experience. When annotating games, annotators ordinarily annotate brilliant moves using exclamation points.
There is a limited amount of narrow algorithmic methods for labeling moves as ``brilliant." For example, a
popular online chess platform, chess.com, classifies ``good piece sacrifices" as brilliant, which not only misses out
on many other types of brilliant moves, but may also be
prone to falsely classifying a move as brilliant~\cite{ChessMoveClassification}. 

% %There
% exists an unexplored opportunity to apply machine learning
% to detect brilliant moves in chess games, thus the objective
% of this research is to develop a classifier that achieves high
% accuracy on categorizing if a given move is brilliant or not.
% Such research would help fill the current gap in our
% understanding of human brilliancy, creativity, and aesthetics in chess, and might reveal insights into our appreciation of
% brilliance or intelligence beyond the domain of chess.

In this paper, we show that it is possible to use the shape of the game tree and the output of chess engines to predict which moves humans would label as brilliant. We collect annotation data from humans from \url{lichess.org}, extract features from the local game tree and the predictions of chess engines, and learn a neural-network-based classifier to use the features to predict whether humans would perceive the move as brilliant. We report that we can obtain an accuracy of 79\% (with 50\% base-rate), a PPV of 83\%, and an NPV of 75\% when classifying moves as perceived as brilliant or not. Interestingly, our analysis shows that our neural network predicts a move to be more brilliant if, holding the rating by a stronger chess engine constant, the weaker chess engine predicts the move to be less good, indicating that part of the perception of brilliance lies in the move not just being good, but also being non-obvious.

Our main intuition is that the game tree provides insight into whether the move is brilliant: a game tree where there are many possibilities for winning indicates that none of the moves would be considered particularly creative or brilliant, whereas a game tree where there is only a single and long path to victory may indicate that finding that path would be considered creative and/or brilliant.

\section{Literature Review}
\subsection{Artificial Intelligence in Chess}
Currently one of the top chess engines is Leela Chess Zero (lc0)~\cite{LCZeroRepo}. This engine uses a Monte Carlo tree
search (MCTS) to expand the game tree. MCTS is
guided by a neural network, which takes in a game state as
input and outputs a vector of move probabilities and a scalar
evaluation estimating the probability of winning from the
game state. The body of the neural network is a residual tower with Squeeze-and-Excitation blocks~\cite{LCZeroNNTopology}. At the time of writing this, lc0 has played over
1.6 billion games against itself, learning through self-play~\cite{LCZero}.

Despite relatively limited research towards human-level play, there is one computer chess engine developed for playing similar to humans: Maia~\cite{McIlroy-Young2020AligningSuperhumanAI}. Maia uses the same architecture as  Leela Chess Zero, however instead of training on self-play, the weights for the neural network guiding the tree search are instead trained on human-played games within fixed bins of player skill ratings. The authors found that Maia can predict the move played by a human more accurately than lc0 and can also predict when a player may make a blunder based on the game state~\cite{McIlroy-Young2020AligningSuperhumanAI}. 

While Maia does model human behaviour in chess more accurately than Stockfish~\cite{Stockfish} or lc0, none of these engines explicitly consider the aesthetic appreciation of chess. Despite the concept of aesthetics seeming vague and challenging to assign a numerical value to, there has been work done towards understanding chess aesthetics and developing a metric correlated to aesthetics. 

\subsection{Chess Aesthetics}
Existing research in chess aesthetics shows the existence and importance of aesthetics in chess~\cite{Margulies1977PrinciplesBeauty}~\cite{Walls1997BeautifulMates}~\cite{LevittFriedgood1995SecretsSpectacularChess}~\cite{Iqbal2006AestheticsChess}, mostly focusing on puzzles. Previous work is either not directly operationalizable or proposes very simple heuristics. In contrast, we propose analyzing the shape of the game tree, which is only made possible through the combination of using a chess engine and a learned model.

%however they have several limitations. So far, the focus has been on checkmate scenarios and puzzles, but aesthetic moves often occur during actual games and well before checkmate occurs. Furthermore, existing computational models for chess aesthetics are very simple, and only consider a subsect of the principles and elements of chess aesthetics. For example, concepts such as depth of analysis or confounding variations are not considered, since they require analysis of the game tree and are much more challenging to compute.
%Hence, there exists an opportunity to address the gaps in both current computer chess engines and our understanding of chess aesthetics by combining the two fields. This can introduce the concept of aesthetics to engines and improve the complexity of models of chess aesthetics through neural networks. Furthermore, by extending aesthetic chess moves to encompass brilliant moves rather than just puzzles, chess aesthetics can be evaluated for a much greater diversity of game states.

\section{Methods}

%The problem can be solved through supervised learning to train the classifier model. As such, the methodology can be split into 3 steps: data collection, feature engineering and model training and evaluation.

In this section, we describe our system for classifying moves as brilliant or not. We collect human-annotated data, extract features from the games, and learn a model that predicts move brilliance.

\subsection{Data Collection \label{sec:data_collection}}
We collect a set of games annotated by users from \url{lichess.org}'s~\cite{LichessAbout} Study feature. The top 624 most popular studies from Lichess were obtained on November 13, 2023.  From these, 8574 total games were extracted. 820 moves were labelled by users as brilliant (two exclamation points) across 556 games, which were part of 158 studies. 1637 moves were labelled as ``good" (one exclamation point). 4518 moves were labelled as other (an annotation is that's not ``bad" is present).

%Additional filters were applied to non-brilliant moves to ensure the accuracy of the classifier is not too inflated by trivially non-brilliant moves. This was done primarily in two ways. First, 1637 ``good" moves were selected from the same 556 games containing brilliant moves. Good moves are a label assigned by the study author, signifying a strong move that reflects positively on the player’s skill. They are lesser versions of brilliant moves in that they share the same characteristics but are not as highly regarded for their intellectual aesthetics. Due to this similarity, the classifier may need to make more complicated inferences to differentiate between brilliant and good moves, and hence are useful to have in the dataset.

%A third category of moves called ``other" moves were selected from the same 556 games as the brilliant moves. These moves were labelled with an annotation that wasn’t explicitly negative for the person making the move. Negatively viewed moves were excluded since they are relatively easy to differentiate from brilliant moves. Unlabelled moves were also excluded from the dataset, since they may be brilliant moves that were skipped over. In total, 4518 other moves were found, so a subset of 1600 other moves were selected randomly to reduce class imbalance and computational load. The good moves and the other moves are both subsets of non-brilliant moves for the purpose of binary classification.

\subsection{Feature Engineering \label{sec:features}}
We compute features from the output of two chess engines. We consider both the evaluation of the moves by engines and the game tree that the engines generate. The engines generate a game tree by exploring mostly promising moves.

We use two engines: lc0~\cite{LCZero}, a strong chess engine trained with self-play (the largest network, T82-768x15x24h-swa-7464000~\cite{LCZero} is used); and Maia, an engine trained to output human-like moves, with an ELO of about 1900~\cite{McIlroy-Young2020AligningSuperhumanAI}~\cite{MaiaWeights2024} (which is quite modest).

For each move, a total of 10 search trees were generated from the board before the move (henceforth called the parent board): once for each of the two weight options (self-play and Maia), and once for each number of nodes evaluated ($10^1$, $10^2$, $10^3$, $10^4$, and $10^5$).

For each tree generated, the following two features can be calculated: is the move of interest in the tree, and if so, did the engine evaluate it as the best move (the move with the highest win probability from the parent board). These features, and specifically the evolution of these features as more nodes are added, may give insight both into the strength of a move and the difficulty of finding it. Both of these attributes reflect on the skill of the player of the move, hence they may be important features for identifying brilliant moves.

We also calculate features for each subtree. A subtree is defined as a tree rooted at a node within the original tree. There are 6 types of subtrees that may be informative for identifying brilliant moves. First, the tree rooted at the parent board gives information about the player’s situation before the move, and the possible moves that could have been made. Second, the tree rooted at the node immediately after the move of interest gives information about the situation following the move, which can give insight into how this move affected the game’s trajectory. The remaining 4 types of subtrees are increasing moves, advantage moves, decreasing moves and disadvantage moves. These are defined based on either the move’s predicted win percentage $Q_{move}$, and the difference between the subtree’s root predicted win percentage and the win percentage of the move: $\Delta Q=Q_{move}-Q_{root}$. Increasing moves have $\Delta Q>0$, advantage moves have $Q_{move}>0$, decreasing moves have $\Delta Q \leq 0$, and disadvantage moves have $Q_{move} \leq 0$. %The move types provide useful information since they help contextualize the strength and difficulty of brilliant moves. If there are no or limited increasing moves other than the move of interest, and/or if there are many decreasing moves or disadvantage moves, the move of interest may appear relatively stronger and more difficult to find, and hence more brilliant.

Since the set of moves belonging to these four types is variable, the features are aggregated so the number of features per full tree (and hence per move that is being classified) is constant. To do this, features are calculated for the subtrees rooted at each increasing, advantage, decreasing and disadvantage move possible from the parent board, excluding the move of interest. The features for each move in each type is aggregated using mean, standard deviation, min and max. This yields $4\times 4\times N$ features per tree from these four types, where N is the number of features per subtree. Combined with the $2\times N$ features from subtrees rooted at the parent board and move of interest, and the 2 binary features calculated per full tree, there are a total of $18\times N+2$ features per full tree. The $N$ features per subtree are as follows: Number of increasing, advantage, decreasing, and disadvantage moves searched from the root node (4 features), predicted win chance at the root node, maximum predicted win chance across all possible moves from the root, probability of search and number of times the root node was searched (2 features), maximum search probability and number of times node was searched across all possible moves from the root (2 features), branching factor of subtree, width at layers 1-7 (7 features), mean, standard deviation and max width across all layers (3 features), and height of the subtree.
In total, this gives $N=22$ features per subtree, or 398 features per tree. Each parent board/move of interest pair in the dataset each had 10 trees generated to cover the 5 node counts ($10^1$ to $10^5$), and the 2 weight options (self-play and Maia). So, each datum has a total of 3980 features. Note, if the tree didn’t contain the move of interest, or if there was no move within one of the move type subsets, a constant vector for the 22 features was used instead. The constant vector contained 0 for all features except for the win chance and max win chance features, which used a value of -1 instead.

\begin{figure}
    \includegraphics[width=9cm]{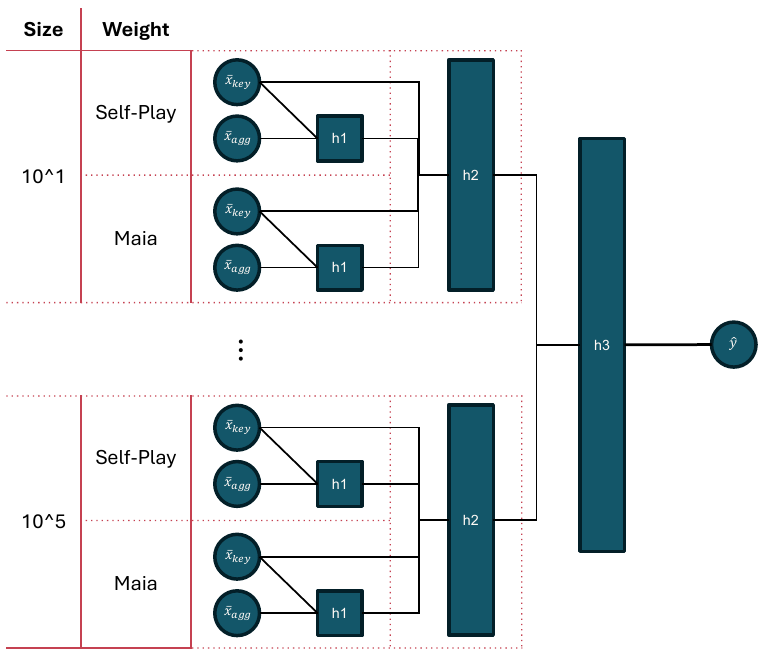}
    \caption{AggReduce architecture. Features for each tree generated with a specific size and weight are processed individually in the first hidden layer, then they are aggregated in tiers.\label{NN}}
\end{figure}

\begin{figure*}[h]
\begin{tabular}{lll}
    \includegraphics[width=5cm]{"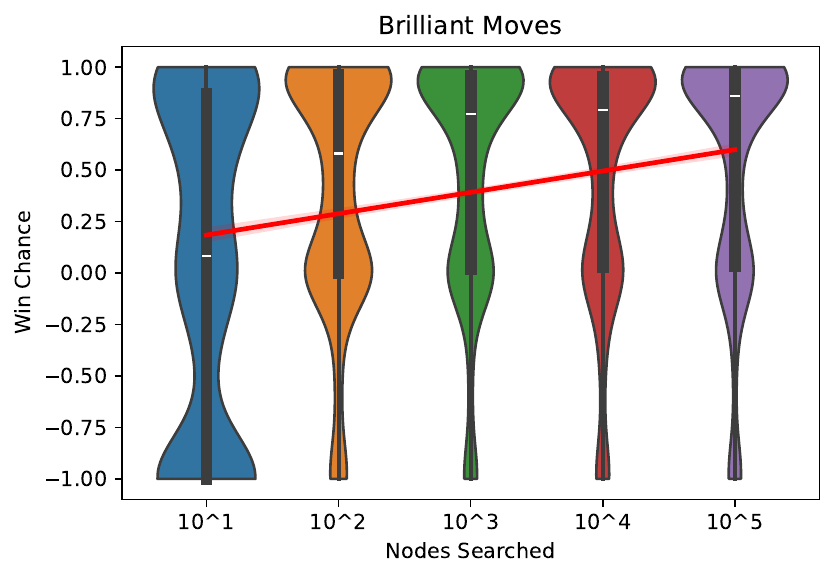"}   & \includegraphics[width=5cm]{"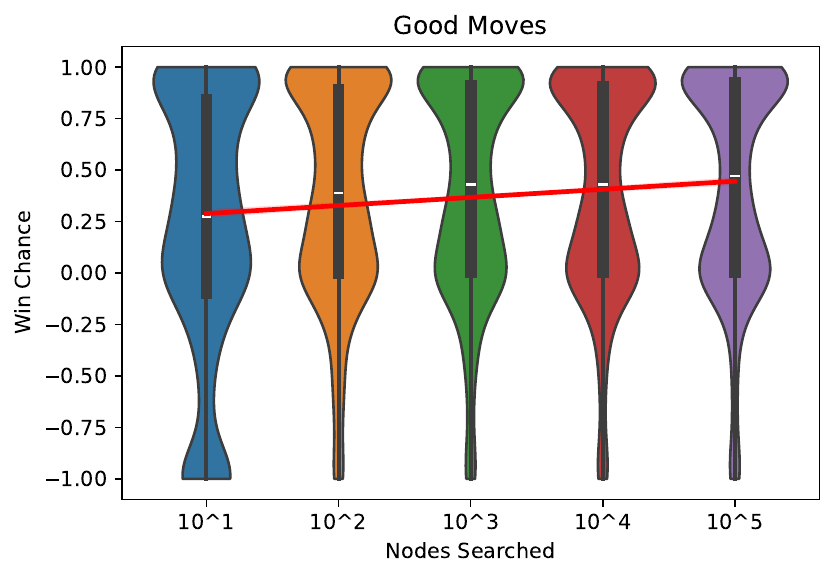"} &
    \includegraphics[width=5cm]{"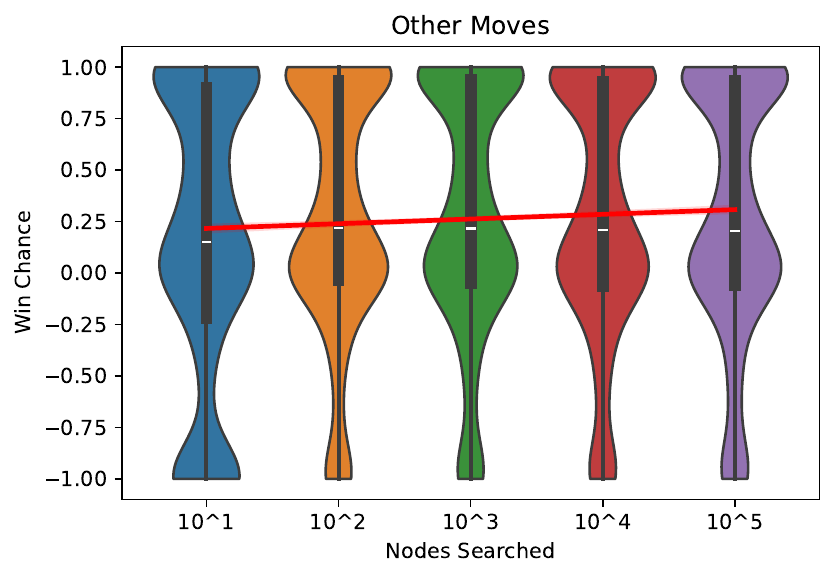"} \\ \includegraphics[width=5cm]{"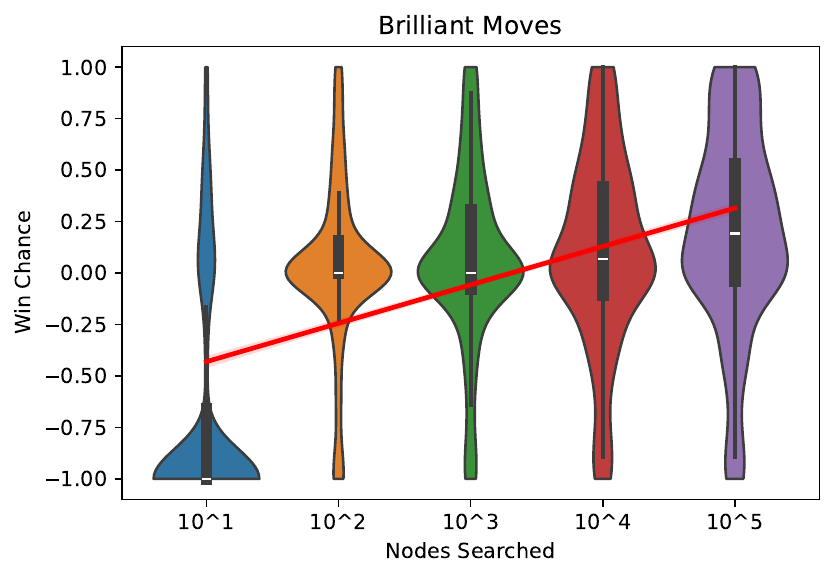"}   & \includegraphics[width=5cm]{"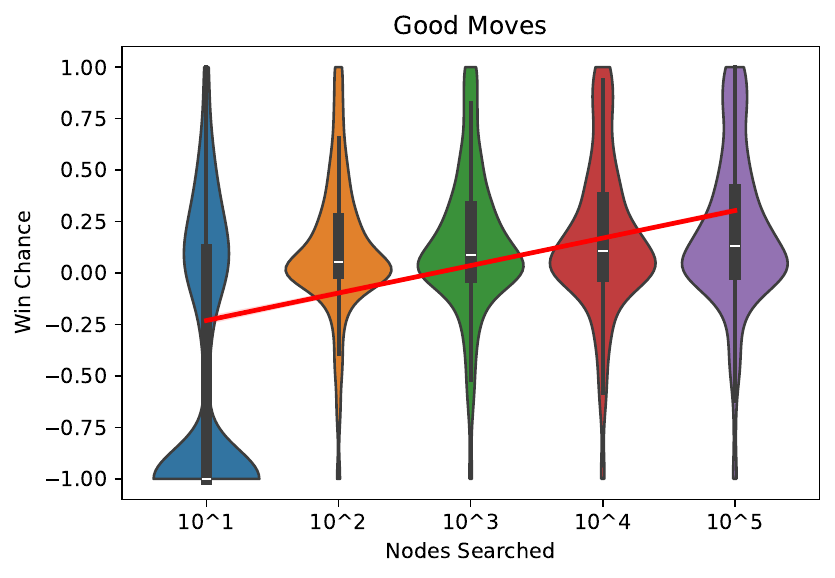"} &
    \includegraphics[width=5cm]{"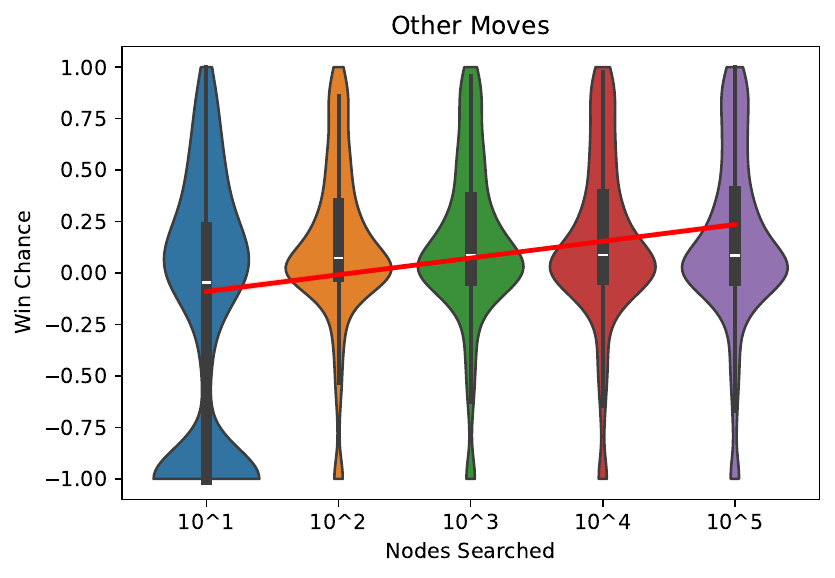"}
\end{tabular}

    \caption{Win chance using self-play weights (top) and Maia weights (bottom) for each move type vs number of nodes searched.  \label{violin}}
\end{figure*}

\subsection{Classifier Models \label{sec:ml}}

A variety of classifier models were trained and tested on the data. This included: logistic regression, a random forest classifier, Gaussian Naïve Bayes, k-nearest neighbors regression, SVM classifier, a fully-connected neural network (one hidden layer with 50 units) and 4 tiered neural networks.

%The motivation of the neural network is to address the feature imbalance caused by the aggregate features for the four move type subtrees.

%These features total to $4\times 4\times 22=352$ per tree, which is much greater than the remaining $2\times 22+2=46$ features. Despite this, the 46 features are likely more informative, since they encode information about the move itself and the overall game state, while the 352 features for the other moves are primarily meant to contextualize the move of interest.

The tiered feedforward neural network architectures operate on the basis that for each input board, a search tree is generated once per weight and per tree size. As there are two weight options and five tree size options, there are ten trees generated per board. Instead of inputting the features for all ten trees at once, tree features can be processed either individually, combined across both weight options, or combined across all the tree sizes. Then, subsequent hidden layers can combine features across the categories that were not yet considered.
Specifically, four tiered architectures were tested: PerWeight, PerSize, PerWeightPerSize and AggReduce. PerWeight’s first hidden layer takes as input the features across all 5 tree sizes for a specific weight. The second hidden layer takes the first layer outputs for both weights, followed by a single output node to classify the move. PerSize follows a similar structure, but the first hidden layer instead takes as input the features across both weights for a specific tree size, then the second hidden layer takes as input the first layer outputs for all 5 tree sizes. PerWeightPerSize adds an additional tier, such that the first hidden layer takes as input the features for a tree generated using a specific weight and tree size. Then, the second hidden layer takes as input the first layer outputs across both weights (still for a specific tree size), and the third hidden layer takes the second layer outputs across all 5 tree sizes.

AggReduce applies the tiered principle, while also addressing the imbalance of the tree features. Recall that each tree has the following features: 2 binary features (is the move of interest in the tree, and is it the best move),  and 22 features for each of the 6 types of subtrees: parent, child, increasing, advantage, decreasing and disadvantage. The two binary features, along with the features for the parent and child subtrees, directly characterize the game state before and after the move of interest. The features for the remaining 4 types of subtrees characterize hypothetical scenarios about what may occur if a different move was taken instead of the move of interest. Since the parent tree already contains information about these moves to some degree, the information gained from these 4 types of subtrees may be less informative for classifying the move. However, since each of the 4 types of subtrees are aggregated using 4 separate techniques (min, max, mean, std), they account for $22\times4\times4=352$ of the total tree features ($\bar{x}_{agg}$). The parent and child subtrees are not aggregated so they each only contribute 22 features, which together with the binary features total to $22\times2+2=46$ key features ($\bar{x}_{key}$). This architecture maps the aggregate features to a smaller hidden layer, allowing the network to extract key information from the aggregate features while focusing on the key features.

The first hidden layer takes as input all the features for a tree generated by a specific weight and tree size; both $\bar{x}_{key}$ and $\bar{x}_{agg}$. This layer has the capability to learn key information from $\bar{x}_{agg}$, contextualized by $\bar{x}_{key}$. The second hidden layer takes the first hidden layers from trees generated across both weight options (still for a specific tree size), along with $\bar{x}_{key}$ for these two trees. Finally, the third hidden layer takes the second hidden layers generated across all 5 tree sizes, followed by a single output node to classify the move. This architecture is shown in Figure~\ref{NN}.

For all neural networks, Each linear layer except for the last applies a 20\% dropout to prevent overfitting and ReLU activation applied after, and the output has a sigmoid activation applied to constrain the classification between 0 and 1. The networks were trained using Adam optimizer, with a learning rate of 1e-4, weight decay of 1e-5, and early stopping with a patience of 10 epochs.

The data was split, withholding 10\% of randomly sampled data for testing. The training set contained 744 brilliant moves, 1463 good moves, and 1444 other moves, while the test set contained 76 brilliant moves, 174 good moves, and 156 other moves. 5-fold cross-validation was used to train and validate each classifier on the training data; the model hyperparameter and final selection was done based on the class-balanced accuracy averaged across each fold. Hyperparameters were selected via grid search.

To validate the significance of the tree features, logistic regression models were trained on various subsets of the data. These subsets are as follows:
\begin{itemize}
    \item Self-play/Maia/Both Win Chance: Using only win chance evaluation before and after the move (and the difference), evaluated with self-play/Maia/both weights and $10^5$ nodes searched.

    \item Self-play/Maia/Both Win Chance per Number of Nodes: Using the win chance evaluation before and after the move, evaluated with self-play/Maia/both weights, for $10^1$,$10^2$,$10^3$,$10^4$, and $10^5$ nodes searched.

    \item Is Best Move: Using only the feature encoding if the move of interest is the best move evaluated, evaluated for both weights and all 5 tree sizes (nodes searched).
\end{itemize}

\section{Results and Discussion \label{sec:results}}
\subsection{Data Analysis}
To verify the utility and information gained by performing the tree search with different nodes searched, the distribution of move win percentages across different number of nodes searched using self-play weights is shown in Fig.~\ref{violin}

These figures show that brilliant moves tend to start off with lower win chance predictions when only 10 nodes are searched, then rises as more nodes are searched. Both good and other moves start off with stronger evaluations with 10 nodes searched, but as more nodes are searched, the brilliant moves peak higher than the non-brilliant moves with a tighter distribution skewed towards a high win chance.

A similar pattern is observed using Maia weights. One key distinction is that brilliant moves appear to be very negatively evaluated when just 10 nodes are searched, which is also the case with good moves, though to less of an effect. Even as more nodes are evaluated, the predicted win chance for brilliant moves remains much lower than those using self-play weights. This supports the idea that brilliant moves are difficult to find, since the human-level chess engine fails to recognize the strength of the move when evaluating a small number of nodes.

% \begin{figure}
% \begin{tabular}{ll}
%     \includegraphics[width=3.5cm]{images/Picture10a.png}   & \includegraphics[width=3.5cm]{images/Picture10b.png} \\
%     \includegraphics[width=3.5cm]{images/Picture10c.png}
% \end{tabular}

%     \caption{Violin plots of win chance using Maia weights for each move type for each number of nodes searched. Linear regression of data points shown in red.
%  \label{violin2}}
% \end{figure}

\subsection{Model Evaluation}

Table~\ref{tab:results} shows the class-balanced accuracies for each model, averaged across 5 folds.

Using only the win chance from the largest tree, the class-balanced accuracy of the logistic regression model only reached 61\%, showing that more features are necessary to characterize brilliant moves more accurately. Varying the number of nodes is informative for classifying brilliant moves; even if the only feature is if the move of interest is the best move, by looking at multiple tree sizes and weights the logistic regression model improved to 64\% accuracy. By using the win chance per number of nodes, the accuracy reaches a maximum of 68\%. Using all the features available, logistic regression reaches an accuracy of 74\%, showing that the full feature set contains additional information useful for classification, aside from just win chance.

The fully connected neural network with one hidden layer of 50 units achieves an average cross validation accuracy of 76\%. Adding a hidden layer using the tiered architecture, both PerWeight and PerSize neural networks achieve an accuracy of 77\%. Adding another hidden layer that first maps features from each individual tree, PerWeightPerSize achieves an accuracy of 78\%. Finally, the AggReduce architecture performed the best, achieving an average cross validation accuracy of 79\% at 21 epochs on average. The hidden layer sizes are 25, 400, 50 for the first, second and third hidden layers respectively. This model was evaluated on the test set to quantify the final accuracy, achieving a test accuracy of 78.60\%.

% Figure~\ref{confusion} shows the confusion matrices for the custom neural network predictions on the test set. We report good prediction

% , normalized by the sample count of each true label to balance the classes. As shown in Figure 11b, the moves labelled as other were very likely to be correctly labelled as non-brilliant, with 93\% of other moves being correctly labelled. Good moves were more likely to be incorrectly labelled, with only 79\% of good moves being correctly labelled as non-brilliant. This is expected, due to the increased similarity between good moves and brilliant moves. 

% \begin{figure}
%     \centering
%     \begin{tabular}{cc}
%         \includegraphics[width=3.5cm]{images/Picture11a.png} & \includegraphics[width=3.5cm]{images/Picture11b.png}          
%     \end{tabular}
    
%     \caption{Confusion matrices of: (a) binary classification true vs predicted labels, on left, and (b) multiclass true labels vs binary predicted labels, on right.
%     \label{confusion}}
% \end{figure}

%Focusing on binary prediction, Fig~\ref{confusion}(a) can be used to calculate the positive and negative predictive values and true positive and negative rates. These values are shown in Table~\ref{tab2}.

When classifying moves as brilliant or not, the AggReduce architecture obtains a true positive rate of 71\%, true negative rate of 85\%, a positive predictive value of 83\% and a negative predictive value of 75\%. In many potential applications of this model, e.g. in automatic game labelling, it may be of particular interest to ensure a move labelled as brilliant is truly brilliant (positive predictive value), and to ensure a move that is truly non-brilliant is labelled as such (true negative rate). %Hence, the relatively higher positive predictive value of 83\% is particularly important.

%  A confusion matrix is shown in Fig.~\ref{confusion}.

% \begin{figure}
% \includegraphics[]{images/Figure 11b.pdf}
% \caption{Confusion matrix when predicting brilliance perception. \label{confusion}}
% \end{figure}

% \begin{table}[]
% \begin{tabular}{|l|l|}
% \hline
% \textbf{} & \textbf{Result} \\ \hline
% PPV & 83\% \\ \hline
% NPV & 75\% \\ \hline
% True Positive Rate & 71\% \\ \hline
% True Negative Rate & 85\% \\ \hline
% \end{tabular}
% \caption{Binary brilliance prediction performance\label{tab2}}
% \end{table}

\subsection{Brilliant moves more often elude the weaker engine \label{sec:maia}}

Further testing was done to better understand the impact of Maia trees on the neural network predictions. For each tree generated with Maia weights, the predicted win chance at the post-move node was set to -1 and the model output score was recalculated. On average, this caused an increase of 0.02554 ($\sigma$ = 0.0471, n = 4057) to the classification score, with a p-value of $<$0.001. This supports the notion that one of the characterising traits of brilliant moves is the difficulty of understanding why they are strong, since reducing Maia’s win chance prediction increased the prediction score of the move being brilliant.

\begin{table}[h]
\begin{tabular}{|l|l|}
\hline
\textbf{Classifier} & \textbf{Accuracy} \\ \hline
\multicolumn{1}{|l|}{LR (Self-play Win Chance)} & \multicolumn{1}{l|}{61\%} \\ \hline
\multicolumn{1}{|l|}{LR (Maia Win Chance)} & \multicolumn{1}{l|}{58\%} \\ \hline
\multicolumn{1}{|l|}{LR (Both Win Chance)} & \multicolumn{1}{l|}{61\%} \\ \hline
\multicolumn{1}{|l|}{LR (Self-play Win Chance per \# of Nodes)} & \multicolumn{1}{l|}{65\%} \\ \hline
\multicolumn{1}{|l|}{LR (Maia Win Chance per \# of Nodes)} & \multicolumn{1}{l|}{65\%} \\ \hline
\multicolumn{1}{|l|}{LR (Both Win Chance per \# of Nodes)} & \multicolumn{1}{l|}{68\%} \\ \hline
\multicolumn{1}{|l|}{LR (Is Best Move)} & \multicolumn{1}{l|}{64\%} \\ \hline
\multicolumn{1}{|l|}{LR (All Features)} & \multicolumn{1}{l|}{74\%} \\ \hline
\multicolumn{1}{|l|}{Random Forest Classifier} & \multicolumn{1}{l|}{74\%} \\ \hline
\multicolumn{1}{|l|}{Gaussian Naïve Bayes} & \multicolumn{1}{l|}{65\%} \\ \hline
\multicolumn{1}{|l|}{K-NN Classifier} & \multicolumn{1}{l|}{66\%} \\ \hline
\multicolumn{1}{|l|}{SVM Classifier} & \multicolumn{1}{l|}{73\%} \\ \hline
\multicolumn{1}{|l|}{Fully-connected NN (50)} & \multicolumn{1}{l|}{76\%} \\ \hline
\multicolumn{1}{|l|}{PerWeight NN (25, 200)} & \multicolumn{1}{l|}{77\%} \\ \hline
\multicolumn{1}{|l|}{PerSize NN (100, 50)} & \multicolumn{1}{l|}{77\%} \\ \hline
\multicolumn{1}{|l|}{PerWeightPerSize NN (100, 100, 25)} & \multicolumn{1}{l|}{78\%} \\ \hline
\multicolumn{1}{|l|}{AggReduce NN (25, 400, 50)} & \multicolumn{1}{l|}{79\%} \\ \hline
\end{tabular}
\caption{5-fold average class-balanced accuracy for each classifier, using the best hyperparameters discovered through cross validation. The best discovered NN hidden layer sizes are displayed in brackets, in order.\label{tab:results}}
\end{table}

\section{Conclusions and Future Work}
In this work, we have shown that it is possible to predict human perception of the brilliance of moves in chess. We have shown that game trees provide information about the perceived brilliance over and above how strong the move is, and that moves that are rated highly by both a weak and a strong engine are predicted to be less brilliant. 

We have studied perception of brilliance by anonymous Lichess users. Expert perception of brilliance might differ from perceptions of Lichess users, and experts and amateurs both can have different tastes within the groups as well.

One can attempt to generalize our work to the perception of brilliance in other domains where creativity has tree-search-like aspects, such as mathematics.

\section{Author Contributions}
KZ and MG planned the study, designed the experiments, and performed the analysis. KZ developed the implementation. MG helped develop the implementation.

\bibliographystyle{iccc}

\bibliography{iccc}

\end{document}